\documentclass[11pt]{article}
\usepackage{acl2014}
\usepackage{times}
\usepackage{url}
\usepackage{latexsym}
\usepackage{amsmath}
\usepackage{amsthm}
\usepackage{amssymb} 
\usepackage{multirow}
\usepackage{mathtools}
\usepackage{graphicx}
\usepackage{subfigure} 
\usepackage{float} 
\usepackage{algorithm}
\usepackage{algorithmic}
\usepackage{caption}
\usepackage{footmisc}
\usepackage{Definitions}
\DeclareTextCommandDefault{\textcopyright}{\textcircled{c}}
\newcommand{\alt}{\textrm{alt}}
\newcommand{\pwr}{\textrm{pwr}}
\newcommand{\plre}{\textrm{plre}}

\usepackage{color}
\newcommand{\ignore}[1]{}

\newcommand{\cii}{c_{i,i-1}}

\newcommand{\diij}{d_{i,i-1}^j}
\newcommand{\ciij}{c_{i,i-1}^{j}}

\newcommand{\ciijsum}{\sum_{i} \ciij}

\newenvironment{itemizesquish}[2]{\begin{list}{\labelitemi}{\setlength{\itemsep}{#1}\setlength{\labelwidth}{#2}\setlength{\leftmargin}{\labelwidth}\addtolength{\leftmargin}{\labelsep}}}{\end{list}}

\title{Language Modeling with Power Low Rank Ensembles}

\author{Ankur P. Parikh \\
  School of Computer Science \\
  Carnegie Mellon University \\
  {\tt apparikh@cs.cmu.edu } \\\And
  Avneesh Saluja \\
Electrical \& Computer Engineering \\
  Carnegie Mellon University \\
  {\tt avneesh@cs.cmu.edu } \\\AND
  Chris Dyer \\
  School of Computer Science \\
  Carnegie Mellon University \\
  {\tt cdyer@cs.cmu.edu } \\\And
  Eric P. Xing \\
  School of Computer Science \\
  Carnegie Mellon University \\
  {\tt epxing@cs.cmu.edu }  \\  }
\date{}

\begin{document}
\maketitle
\begin{abstract}
We present power low rank ensembles (PLRE), a flexible framework for $n$-gram language modeling where ensembles of low rank matrices and tensors are used to obtain smoothed probability estimates of words in context.
Our method can be understood as a generalization of $n$-gram modeling to non-integer $n$, and includes standard techniques such as absolute discounting
and Kneser-Ney smoothing as special cases. PLRE training is efficient and our approach outperforms state-of-the-art modified Kneser Ney baselines in terms of perplexity on large corpora as well as on BLEU score in a downstream machine translation task.
\end{abstract}

\vspace{-2mm}
\section{Introduction}
\label{sec:intro}
Language modeling is the task of estimating the probability of sequences of words in a language and is an important component in,
among other applications, automatic speech recognition \cite{rabiner1993fundamentals} and machine translation \cite{koehn2010}.
The predominant approach to language modeling is the $n$-gram model, wherein the probability of a word sequence $P(w_1, \dots, w_{\ell})$ 
is decomposed using the chain rule, and then a Markov assumption is made:
$
  P(w_1, \dots, w_{\ell})  \approx \prod_{i=1}^{\ell} P(w_i | w_{i-n+1}^{i-1}) .
$
While this assumption substantially reduces the modeling complexity, parameter estimation remains a major challenge. \ignore{ the maximum likelihood estimator performs quite poorly.\ignore{Vocabulary sizes are often significant (often $10^5-10^6$ word types, or larger), leading to inaccurate estimation of joint and conditional counts from the data, even with the web-scale corpora available these days.}} 
Due to the power-law nature of language \cite{Zipf1949}, the maximum likelihood estimator massively overestimates the probability of rare events and assigns zero probability to legitimate word sequences that happen not to have been observed in the training data \cite{manning1999foundations}.
\ignore{
 able to estimate some common subsequences very accurately while impractically
assigning zero probability to rare, but legitimate word sequences that are not observed in the training data.}

Many  smoothing techniques have been proposed to address the estimation challenge. These reassign probability mass (generally from over-estimated events) to unseen word sequences, whose probabilities are estimated by interpolating with or backing off to lower order $n$-gram models \cite{Chen1999}. 


Somewhat surprisingly, these widely used smoothing techniques differ substantially from techniques for coping with data sparsity in other domains, such as collaborative filtering~\cite{koren2009matrix,su2009survey} or matrix completion~\cite{candes2009exact,cai2010singular}. In these areas, \textit{low rank} approaches based on matrix factorization play a central role~\cite{seung2001algorithms,salakhutdinov2008bayesian,mackey2011divide}. For example, in recommender systems, a key challenge is dealing with the sparsity of ratings from a single user, since typical users will have rated only a few items. By projecting the low rank representation of a user's (sparse) preferences into the original space, an estimate of ratings for new items is obtained. These methods are attractive due to their computational efficiency and mathematical well-foundedness.

In this paper, we introduce \textbf{power low rank ensembles} (PLRE), in which low rank tensors are used to produce smoothed estimates for $n$-gram probabilities. Ideally, we would like the low rank structures to discover semantic and syntactic relatedness among words and $n$-grams, which are used to produce smoothed estimates for word sequence probabilities.  
\ignore{In this work, we present a low rank approach for $n$-gram language modeling, called \textbf{power low rank ensembles} (PLRE), where low rank matrices and tensors are used to produce smoothed estimates for word sequence probabilities.} 
In contrast to the few previous low rank language modeling approaches, PLRE is not orthogonal to $n$-gram models, but rather a general framework where existing $n$-gram smoothing methods such as Kneser-Ney smoothing are special cases.  A key insight is that PLRE does not compute low rank approximations of the original joint count matrices (in the case of bigrams) or tensors i.e. multi-way arrays (in the case of 3-grams and above), but instead altered quantities of these counts based on an element-wise power operation, similar to how some smoothing methods modify their lower order distributions.

Moreover, PLRE has two key aspects that lead to easy scalability for large corpora and vocabularies. First, since it utilizes the original $n$-grams, the ranks required for the low rank matrices and tensors tend to be remain tractable (e.g. around $100$ for a vocabulary size $V \approx 1 \times 10^6$) leading to fast training times. This differentiates our approach over other methods  that leverage an underlying latent space such as neural networks~\cite{Bengio2003,mnih2007three,Mikolov2010}  or soft-class models~\cite{saul1997aggregate} where the underlying dimension is required to be quite large to obtain good performance. Moreover, at test time, the probability of a sequence can be queried in time $O(\kappa_{max})$ where $\kappa_{max}$ is the maximum rank of the low rank matrices/tensors used. While this is larger than Kneser Ney's virtually constant query time, it is substantially faster than conditional exponential family models~\cite{chen2000survey,Chen2009,Nelakanti2013} and neural networks which require $O(V)$ for exact computation of the normalization constant.  See Section~\ref{sec:related} for a more detailed discussion of related work.

\textbf{Outline:} We first review existing $n$-gram smoothing methods (\S\ref{sec:background}) and then present the intuition behind the key components of our technique: \textbf{rank} (\S\ref{sec:rank}) and \textbf{power} (\S\ref{sec:power}). We then show how these can be interpolated into an ensemble (\S\ref{sec:ensemble}).  In the experimental evaluation on English and Russian corpora (\S\ref{sec:experiments}), we find that PLRE outperforms Kneser-Ney smoothing and all its variants, as well as class-based language models.  
 We also include a comparison to the log-bilinear neural language model~\cite{mnih2007three} and evaluate performance on a downstream machine translation task (\S\ref{sec:mt}) where our method achieves consistent improvements in BLEU.
\vspace{-2mm}
\section{Discount-based Smoothing}
\label{sec:background}

We first provide background on absolute discounting~\cite{Ney1994} and Kneser-Ney smoothing~\cite{kneser1995improved}, two common $n$-gram smoothing methods. Both methods can be formulated as back-off or interpolated models; we describe the latter here since that is the basis of our low rank approach.

  
\subsection{Notation}
Let $c(w)$ be the count of word $w$, and
similarly $c(w,w_{i-1})$ for the joint count of words $w$ and
$w_{i-1}$. For shorthand we will define $w^{j}_{i}$ to denote the word sequence $\{w_i, w_{i+1},...,w_{j-1}, w_j \}$.  
Let $\Phat(w_i)$ refer to the maximum likelihood estimate (MLE) of
the probability of word $w_i$, and similarly $\Phat(w_i |
w_{i-1})$ for the probability conditioned on a history, or more
generally, $\Phat(w_i | w^{i-1}_{i-n+1})$.

Let  $N_{-}(w_{i}) := |\{ w  : c(w_i, w) > 0\} |$ be the
number of distinct words that appear before $w_{i}$.  
More generally, let $N_{-} (w^{i}_{i-n+1}) =  |\{w : c(w^{i}_{i-n+1}, w) > 0 \} |$.
Similarly, let $N_{+} (w^{i-1}_{i-n+1}) =  |\{w : c(w, w^{i-1}_{i-n+1}) > 0 \} |$. $V$ denotes the vocabulary size.

\subsection{Absolute Discounting}
\label{sec:absolutediscount}
Absolute discounting works on the idea of interpolating higher order
$n$-gram models with lower-order $n$-gram
models. However, first some probability mass must be ``subtracted" from the higher order $n$-grams so that the leftover probability can be allocated to the lower order $n$-grams. More specifically, define the following discounted conditional probability:

\begin{eqnarray}
\Phat_{D}(w_i | w^{i-1}_{i-n+1}) = \frac{\max \{ c(w_i, w^{i-1}_{i-n+1}) - D, 0 \}}{ c(w^{i-1}_{i-n+1})} \nonumber
\end{eqnarray}
Then absolute discounting $P_{\textrm{abs}}(\cdot)$ uses the following (recursive) equation:
\begin{flalign}
& P_{\textrm{abs}}(w_i | w^{i-1}_{i-n+1}) = \Phat_{D}(w_i | w^{i-1}_{i-n+1})  \notag \\
 &\qquad \qquad + \gamma(w^{i-1}_{i-n+1}) P_{\textrm{abs}}(w_i |
 w^{i-1}_{i-n+2}) \nonumber
\end{flalign}
where  $\gamma(w^{i-1}_{i-n+1})$ is the leftover weight (due to the
discounting) that is chosen so that the conditional distribution sums to one: $\gamma(w^{i-1}_{i-n+1}) = \frac{D}{c(w^{i-1}_{i-n+1})}N_{+}(w_{i-n+1}^{i-1})$. 
For the base case, we set $P_{\textrm{abs}}(w_i) = \Phat(w_i)$.  

\noindent \textbf{Discontinuity:} Note that if $c(w^{i-1}_{i-n+1}) = 0$, then $\gamma(w^{i-1}_{i-n+1}) =
\frac{0}{0}$, in which
case $\gamma(w^{i-1}_{i-n+1})$ is set to 1.  We will see that this discontinuity appears in PLRE as well. 

%
%
\subsection{Kneser Ney Smoothing}
Ideally, the smoothed probability should preserve the observed unigram distribution:
\begin{eqnarray}
\Phat(w_i) = \sum_{w^{i-1}_{i-n+1}} P_{\textrm{sm}}(w_i | w^{i-1}_{i-n+1}) \Phat(w^{i-1}_{i-n+1})
\label{eq:margconst}
\end{eqnarray}
where $P_{\textrm{sm}}(w_i |  w^{i-1}_{i-n+1})$ is the smoothed conditional
probability that a model outputs.  
Unfortunately, absolute discounting does
not satisfy this property, since it exclusively uses the unaltered MLE unigram
model as its lower order model.  In practice, the lower order distribution
is only utilized when we are unsure about the higher order
distribution (i.e., when $\gamma(\cdot)$ is large).  Therefore, the
unigram model should be altered to condition on this fact.  

This is the inspiration behind Kneser-Ney (KN) smoothing, an elegant
algorithm with robust performance in $n$-gram language modeling. 
KN smoothing defines alternate probabilities
$P^{\textrm{alt}}(\cdot)$: 
\begin{align*}
\hspace{-0.5cm}  P^{\textrm{alt}}_D(w_i | w^{i-1}_{i-n'+1}) &= \begin{cases} \Phat_{D}(w_i | w^{i-1}_{i-n'+1}), &\hspace{-0.3cm} \mbox{if } n' = n \\ \\
\frac{\max \{ N_{-} (w^{i}_{i-n'+1}) - D, 0 \} }{\sum_{w_i} N_{-}
  (w^{i}_{i-n'+1})}, &\hspace{-0.3cm} \mbox{if } n' < n \end{cases}
\end{align*}

%

The base case for unigrams reduces to $P^{\textrm{alt}}(w_i) = \frac{N_{-}(w_i)}{\sum_{w_i} N_{-}(w_i)}$. Intuitively $P^{\textrm{alt}}(w_i)$ is proportional to the number of unique words that precede $w_i$. Thus, words that appear in many different contexts will be given higher weight than words that consistently appear after only a few contexts.  These alternate distributions are then used with absolute discounting:
\begin{flalign}
& P_{\textrm{kn}}(w_i | w^{i-1}_{i-n+1}) = P_{D}^{\textrm{alt}}(w_i | w^{i-1}_{i-n+1})  \notag \\
 &\qquad \qquad + \gamma(w^{i-1}_{i-n+1}) P_{\textrm{kn}}(w_i | w^{i-1}_{i-n+2})
 \label{eq:kney}
\end{flalign}
where we set $P_{\textrm{kn}}(w_i) = P^{\textrm{alt}}(w_i)$.  By definition, KN smoothing satisfies the marginal constraint in Eq.~\ref{eq:margconst} \cite{kneser1995improved}.

\vspace{-2mm}
\section{Power Low Rank Ensembles}
In $n$-gram smoothing methods, if a bigram count $c(w_i,
w_{i-1})$ is zero, the
unigram probabilities are used, which is equivalent to assuming that
$w_i$ and $w_{i-1}$ are independent ( and similarly for general $n$).  However, in this
situation, instead of backing off to a $1$-gram, we may like to
back off to a ``$1.5$-gram" or more generally an order between 1 and 2
that captures a coarser level of
dependence between $w_i$ and $w_{i-1}$ and does not assume full
independence.  

Inspired by this intuition, our strategy is to construct an ensemble
of matrices and tensors that not only consists of MLE-based count
information, but also contains quantities that represent levels of dependence
in-between the various orders in the model. We call these combinations
power low rank ensembles (PLRE), and they can be thought of as
$n$-gram models with non-integer $n$.  Our approach can be recursively
formulated as: 

\begin{flalign}
& P_{\textrm{plre}}(w_i | w^{i-1}_{i-n+1}) = P_{\Db_0}^{\textrm{alt}}(w_i | w^{i-1}_{i-n+1}) \notag \\
&+ \gamma_0(w^{i-1}_{i-n+1}) \bigg ( \Zb_{\Db_1}(w_i | w^{i-1}_{i-n+1})  + .....  \notag \\
& + \gamma_{\eta-1}(w^{i-1}_{i-n+1}) \bigg  (  \Zb_{\Db_\eta} (w_i | w^{i-1}_{i-n+1})  \notag \\
&+ \gamma_{\eta}(w^{i-1}_{i-n+1}) \bigg (  P_{\textrm{plre}}(w_i | w^{i-1}_{i-n+2}) \bigg  ) \bigg ) ...  \bigg )
\label{eq:overall}
\end{flalign}
where $\Zb_1,...,\Zb_{\eta}$ are conditional probability
matrices that represent the intermediate $n$-gram orders\footnote{with a slight abuse of notation, let $\Zb_{\Db_j}$
  be shorthand for $\Zb_{j, \Db_j}$} and $\Db$ is a discount function
(specified in \S\ref{sec:ensemble}).

This formulation begs answers to a few critical questions. How
to construct matrices that represent conditional probabilities for intermediate $n$?
How to transform them in a way that generalizes the altered lower
order distributions in KN smoothing? How to combine these matrices
such that the marginal constraint in Eq.~\ref{eq:margconst} still
holds?  The following propose solutions to these three
queries:
 \begin{enumerate}
 \item
 \textbf{Rank} (Section~\ref{sec:rank}): \textit{Rank} gives us a
 concrete measurement of the dependence between $w_i$ and $w_{i-1}$.  By
 constructing low rank approximations of the bigram count matrix and
 higher-order count tensors, we
 obtain matrices that represent coarser dependencies, with a rank one
 approximation implying that the variables are independent. 
 \item
 \textbf{Power} (Section~\ref{sec:power}): In KN smoothing, the lower order distributions are not the original counts but rather altered estimates. We propose  a continuous generalization of this alteration by taking the element-wise \textit{power} of the counts.
 \item
 \textbf{Creating the Ensemble} (Section~\ref{sec:ensemble}): Lastly,
 PLRE also defines a way to interpolate the specifically constructed
 intermediate $n$-gram matrices. Unfortunately a constant discount, as
 presented in Section~\ref{sec:background}, will not in general
 preserve the lower order marginal constraint
 (Eq.~\ref{eq:margconst}). We propose a generalized discounting scheme
 to ensure the constraint holds.
 \end{enumerate}

\subsection{Rank}
 \label{sec:rank} 
We first show how rank can be utilized to construct quantities
between an $n$-gram and an $n-1$-gram. In general, we think of an
$n$-gram as an  $n^{\textrm{th}}$ order tensor i.e. a multi-way array with $n$ indices $\{ i_1,...,i_n \}$. (A vector is a tensor of order 1, a matrix is a tensor of order 2 etc.)     Computing a
special rank one approximation of slices of this tensor produces the
$n-1$-gram.  Thus, taking rank $\kappa$ approximations in this fashion
allows us to represent dependencies between
an $n$-gram and $n-1$-gram.  
  
 Consider the bigram count matrix $\Bb$ with $N$ counts which has rank $V$. Note that $\Phat(w_i | w_{i-1}) = \frac{\Bb(w_i, w_{i-1})}{\sum_{w} \Bb(w, w_{i-1})}$.
 Additionally, $\Bb$ can be considered a random variable that is the
 result of sampling $N$ tuples of $(w_i, w_{i-1})$ and agglomerating
 them into a count matrix.  Assuming $w_i$ and $w_{i-1}$ are independent, the expected value (with respect to the empirical distribution)
$\EE[\Bb] = N P(w_i) P(w_{i-1})$, which can be rewritten as being
proportional to the outer product of the unigram probability vector
with itself, and is thus rank one.  

This observation extends to higher order $n$-grams as well. Let
$\Cb^n$ be the $n^{\textrm{th}}$ order tensor where
$\Cb^n(w_i,....,w_{i-n+1}) = c(w_i,...,w_{i-n+1})$. Furthermore denote
$\Cb^n(:,\tilde{w}^{i-1}_{i-n+2},:)$ to be the $V \times V$ matrix slice
of $\Cb^n$ where $w_{i-n+2},...,w_{i-1}$ are held fixed to a particular sequence
$\tilde{w}_{i-n+2},...,\tilde{w}_{i-1}$. Then if $w_i$ is conditionally
independent of $w_{i-n+1}$ given $w^{i-1}_{i-n+2}$, then $\EE[\Cb^n(:,\tilde{w}^{i-1}_{i-n+2},:)]$ is rank one $\forall \tilde{w}^{i-1}_{i-n+2}$.

 
 

However, it is rare that these matrices are actually rank one, either
due to sampling variance or the fact that $w_i$ and $w_{i-1}$ are not independent. What we
would really like to say is that the \textit{best} rank one
approximation $\Bb^{(1)}$ (under some norm) of $\Bb$ is $\propto
\Phat(w_i) \Phat(w_{i-1})$.  While this statement is not true under
the $\ell_2$ norm, it is true under generalized KL divergence~\cite{seung2001algorithms}:
$gKL(\Ab || \Bb) = \sum_{ij} \left ( \Ab_{ij} \log (\frac{\Ab_{ij}}{\Bb_{ij}}) - \Ab_{ij} + \Bb_{ij} ) \right )$. 

In particular, generalized KL divergence preserves row and column sums: \textit{if $\Mb^{(\kappa)}$ is the best rank $\kappa$
  approximation of $\Mb$ under $gKL$ then the row sums and
  column sums of $\Mb^{(\kappa)}$ and $\Mb$ are
  equal}~\cite{ho2008non}.  Leveraging this property, it is straightforward to prove the following lemma:
\begin{lemma}
Let $\Bb^{(\kappa)}$ be the best rank $\kappa$ approximation of $\Bb$ under gKL. Then $\Bb^{(1)} \propto \Phat(w_i) \Phat(w_{i-1})$ and 
$\forall w_{i-1}$ s.t. $c(w_{i-1}) \neq 0$: 
\begin{eqnarray}
\Phat(w_i) = \frac{\Bb^{(1)}(w_i, w_{i-1})}{\sum_{w} \Bb^{(1)}(w, w_{i-1})} \nonumber
\end{eqnarray}

For more general $n$, let $\Cb^{n,(\kappa)}_{i-1,...,i-n+2}$ be the best rank $\kappa$ approximation of $\Cb^n(:,\tilde{w}^{i-1}_{i-n+2},:)$ under $gKL$. Then similarly,
$\forall w^{i-1}_{i-n+1}$ s.t. $c(w^{i-1}_{i-n+1}) > 0$:
\begin{flalign}
&\Phat(w_i | w_{i-1},...,w_{i-n+2}) \notag \\
& = \frac{\Cb^{n, (1)}_{i-1,...,i-n+2}(w_i, w^{i-1}_{i-n+1})}{\sum_{w} \Cb^{n, (1)}_{i-1,...,i-n+2}(w,w^{i-1}_{i-n+1})} 
\end{flalign}
\label{lem:rank}
\end{lemma}
Thus, by selecting $1 < \kappa < V$, we obtain count matrices and
tensors between $n$ and $n-1$-grams. The condition that $c(w_{i-n+1}^{i-1}) > 0$ corresponds to
the discontinuity discussed in \S\ref{sec:absolutediscount}.  
%
%
%
%
%

\subsection{Power}
\label{sec:power}
Since KN smoothing alters the lower order distributions instead of
simply using the MLE, varying the rank is not sufficient in order to
generalize this suite of techniques.  Thus, PLRE computes low rank approximations of altered count matrices.
Consider taking the elementwise power $\rho$ of the bigram count
matrix, which is denoted by $\Bb^{\cdot \rho}$. For example, the observed bigram count matrix and associated row sum:
\begin{eqnarray}
\footnotesize
\Bb^{\cdot 1} = \left( \begin{array}{ccc}
1.0 & 2.0 & 1.0 \\
0 & 5.0 & 0 \\
2.0 & 0 & 0
\end{array} \right)
 \stackrel{\textrm{row sum}}{\rightarrow} 
\left( \begin{array}{c}
4.0  \\
5.0 \\
2.0
\end{array} \right) \nonumber
\end{eqnarray}

As expected the row sum is equal to the unigram counts (which we
denote as $\ub$).  Now consider $\Bb^{\cdot 0.5}$:
\begin{eqnarray}
\footnotesize
\Bb^{\cdot 0.5} = \left( \begin{array}{ccc}
1.0 & 1.4 & 1.0 \\
0 & 2.2 & 0 \\
1.4 & 0 & 0
\end{array} \right)
 \stackrel{\textrm{row sum}}{\rightarrow} 
\left( \begin{array}{c}
3.4  \\
2.2 \\
1.4
\end{array} \right) \nonumber
\end{eqnarray}
Note how the row sum vector has been altered. In particular since $w_1$ (corresponding to the first row) has a more diverse
history than $w_2$, it has a higher row sum (compared to in $\ub$
where $w_2$ has the higher row sum). 
 Lastly, consider the case when $p = 0$:
\begin{eqnarray}
\footnotesize
\Bb^{\cdot 0} = \left( \begin{array}{ccc}
1.0 & 1.0 & 1.0 \\
0 & 1.0 & 0 \\
1.0 & 0 & 0
\end{array} \right)
 \stackrel{\textrm{row sum}}{\rightarrow} 
\left( \begin{array}{c}
3.0  \\
1.0 \\
1.0
\end{array} \right) \nonumber
\end{eqnarray}
The row sum is now the number of unique words that precede $w_i$ (since $\Bb^{0}$ is binary) and is thus equal to the (unnormalized) Kneser Ney unigram. This idea also generalizes to higher order $n$-grams and leads us to the following lemma:
\begin{lemma}
Let $\Bb^{(\rho, \kappa)}$ be the best rank $\kappa$ approximation of $\Bb^{\cdot \rho}$ under gKL. Then $\forall w_{i-1}$ s.t. $c(w_{i-1}) \neq 0$: 
\begin{eqnarray}
P^{\textrm{alt}}(w_i) = \frac{\Bb^{(0, 1)}(w_i, w_{i-1})}{\sum_{w} \Bb^{(0, 1)}(w, w_{i-1})} \nonumber
\end{eqnarray}
For more general $n$, let $\Cb^{n,(\rho, \kappa)}_{i-1,...,i-n+2}$ be
the best rank $\kappa$ approximation of $\Cb^{n,
  (\rho)}(:,\tilde{w}^{i-1}_{i-n+2},:)$ under $gKL$.  Similarly,
$\forall w^{i-1}_{i-n+1}$ s.t. $c(w^{i-1}_{i-n+1}) > 0$:
\begin{flalign}
&P^{\textrm{alt}}(w_i | w_{i-1},...,w_{i-n+2}) \notag \\
& = \frac{\Cb^{n, (0, 1)}_{i-1,...,i-n+2}(w_i, w^{i-1}_{i-n+1})}{\sum_{w} \Cb^{n, (0, 1)}_{i-1,...,i-n+2}(w,w^{i-1}_{i-n+1})} 
\end{flalign}
\label{lem:power}
\end{lemma}

\vspace{-2mm}
\section{Creating the Ensemble}
\label{sec:ensemble}
Recall our overall formulation in Eq.~\ref{eq:overall}; a naive
solution would be to set $\Zb_1,...,\Zb_{\eta}$  to low rank
approximations of the count matrices/tensors under varying powers, and
then interpolate through constant absolute discounting. Unfortunately, the
marginal constraint in Eq.~\ref{eq:margconst} will generally not hold
if this strategy is used.  Therefore, we propose a generalized discounting scheme
where each non-zero $n$-gram count is associated with a different
discount $\Db_j(w_i, w^{i-1}_{i-n'+1})$. The low rank approximations are then computed on the
discounted matrices, leaving the marginal constraint intact. 

For clarity of exposition, we focus on the special case where $n = 2$ with only one low rank matrix before stating our general algorithm:
\begin{flalign}
& P_{\textrm{plre}}(w_i | w_{i-1}) = \Phat_{\Db_0}(w_i | w_{i-1}) \notag \\
&+ \gamma_0(w_{i-1}) \bigg ( \Zb_{\Db_1}(w_i | w_{i-1})   + \gamma_{1}(w_{i-1}) P^{\alt}(w_i) \bigg )
\label{eq:overall_bigram}
\end{flalign}

Our goal is to compute $\Db_{0}, \Db_1$ and $\Zb_1$  so that the following lower order marginal constraint holds:
\begin{flalign}
\Phat(w_{i}) = \sum_{w_{i-1}} P_{\plre}(w_i | w_{i-1}) \Phat(w_{i-1})
\label{eq:intermargconst}
\end{flalign}

Our solution can be thought of as a two-step procedure where we
compute the discounts $\Db_{0}, \Db_1$ (and the
$\gamma(w_{i-1})$ weights as a by-product), followed by the low rank quantity
$\Zb_1$.  First, we construct the following intermediate ensemble of powered, but full rank terms. Let $\Yb^{\rho_j}$ be the matrix such that $\Yb^{\rho_j}(w_i, w_{i-1}) := c(w_i, w_{i-1})^{\rho_j}$. Then define 
\begin{flalign}
& P_{\pwr}(w_i | w_{i-1}) := \Yb_{\Db_0}^{(\rho_0 = 1)}(w_i | w_{i-1}) \notag \\
&\quad + \gamma_0(w_{i-1}) \bigg ( \Yb_{\Db_1}^{(\rho_1)} (w_i | w_{i-1}) \notag \\
&\quad +  \gamma_{1}(w_{i-1})\Yb^{(\rho_{2} = 0)}
(w_i | w_{i-1}) \bigg )  
\end{flalign} 
where with a little abuse of notation:
{\footnotesize
\begin{eqnarray*}
 \Yb_{\Db_j}^{\rho_j}(w_i | w_{i-1}) = \frac{c(w_i, w_{i-1})^{\rho_j} - \Db_j(w_i, w_{i-1})}{\sum_{w_i}c(w_i, w_{i-1})^{\rho_j}}
\end{eqnarray*}}
Note that $P^{\textrm{alt}}(w_i)$
has been replaced with
$\Yb^{(\rho_{2} = 0)}  (w_i | w_{i-1})$, based on Lemma~\ref{lem:power}, and will equal $P^{\textrm{alt}}(w_i)$ once the low rank approximation is taken as discussed in \S~\ref{sec:lrapprox}).  

Since we have only combined terms of different power (but all full
rank), it is natural choose the discounts so that the result remains
unchanged i.e., $P_{\textrm{pwr}}(w_i | w_{i-1}) =
\Phat(w_i | w_{i-1})$, since the low rank approximation (not the power) will implement smoothing. Enforcing this
constraint gives rise to a set of linear equations that can be solved (in closed form) to obtain the discounts as we now show below.

\subsection{Step 1: Computing the Discounts}
\label{sec:computedisc}
To ensure the constraint that $P_{\textrm{pwr}}(w_i | w_{i-1}) =
\Phat(w_i | w_{i-1})$, it is sufficient to enforce the following two local constraints:
\begin{flalign}
 & \Yb^{(\rho_j)} (w_i | w_{i-1})  =  \Yb_{\Db_j}^{(\rho_j)} (w_i | w_{i-1}) \notag \\ 
 &\quad + \gamma_j(w_{i-1}) \Yb^{(\rho_{j+1})} (w_i | w_{i-1}) \textrm{ for } j=0,1 
 \label{eq:locald}
\end{flalign}
This allows each $\Db_j$ to be solved for independently of the other
$\{ \Db_{j'} \}_{j' \neq j}$.  Let $\cii = c(w_i, w_{i-1})$, $\ciij = c(w_{i}, w_{i-1})^{\rho_j}$,  and $\diij = \Db_j(w_{i}, w_{i-1})$.  Expanding
Eq.~\ref{eq:locald} yields that $\forall w_i, w_{i-1}$:
\begin{flalign}
 & \frac{\ciij}{\ciijsum} = \notag \\
 & \frac{\ciij- \diij}{\ciijsum} + \left ( \frac{\sum_{i} \diij}{\ciijsum} \right ) \frac{\cii^{j+1}}{\sum_{i} \cii^{j+1}} 
\end{flalign}
which can be rewritten as:
\begin{flalign}
- \diij+  \left ( \sum_{i} \diij \right ) \frac{\cii^{j+1}}{\sum_{i} \cii^{j+1}}  = 0
\label{eq:system}
\end{flalign}
Note that Eq.~\ref{eq:system} decouples across $w_{i-1}$
since the only $\diij$ terms that are dependent are the ones that
share the preceding context $w_{i-1}$.

It is straightforward to see that setting $\diij$ proportional to $\cii^{j+1}$ satisfies Eq.~\ref{eq:system}. 
Furthermore it can be shown that all solutions are of this form (i.e., the linear system has a null space of exactly one). Moreover, we are interested in a particular subset of solutions where a single parameter $d_{\ast}$ (independent of $w_{i-1}$) controls the scaling as indicated by the following lemma:

\begin{lemma}
Assume that $\rho_j \geq \rho_{j+1}$. Choose any $0 \leq d_{\ast} \leq
1$. Set $\diij = d_\ast \cii^{j+1} \, \forall i,j $. The resulting discounts satisfy Eq.~\ref{eq:system}
as well as the inequality constraints $0 \leq \diij \leq \ciij$. Furthermore, the leftover weight $\gamma_j$ takes the form:
\begin{eqnarray}
\gamma_j(w_{i-1}) = \frac{\sum_{i} \diij }{\ciijsum}  = \frac{d_{\ast} {\sum_{i} \cii^{j+1}} }{\ciijsum} \notag
\end{eqnarray}
\begin{proof}
Clearly this choice of $\diij$ satisfies Eq.~\ref{eq:system}. The largest possible value of $\diij$ is $\cii^{j+1}$. $\rho_j \geq \rho_{j+1}$, implies $\ciij \geq \cii^{j+1}$. Thus the inequality constraints are met. It is then easy to verify that $\gamma$ takes the above form.
\end{proof}
\label{lem:disc}
\end{lemma}
The above lemma generalizes to longer contexts (i.e. $n > 2$) as shown in Algorithm~\ref{alg:computed}. Note that if $\rho_{j} = \rho_{j+1}$ then
Algorithm~\ref{alg:computed} is equivalent to scaling the counts e.g. deleted-interpolation/Jelinek Mercer smoothing~\cite{jelinek1980interpolated}.
On the other hand, when $\rho_{j+1} = 0$, Algorithm~\ref{alg:computed} is equal to the absolute discounting that is used in Kneser-Ney.
Thus, depending on $\rho_{j+1}$, our method generalizes different types of interpolation schemes to construct an ensemble so that the marginal constraint
is satisfied.

\subsection{Step 2: Computing Low Rank Quantities}
\label{sec:lrapprox}

\begin{algorithm}[t!]
\caption{\small Compute $\Db$}
\textbf{In}: Count tensor $\Cb^{n}$, powers $\rho_j, \rho_{j+1}$ such that $\rho_j \geq \rho_{j+1}$, and parameter $d_\ast$. \\
\setlength{\abovedisplayskip}{2pt}
\setlength{\abovedisplayshortskip}{0pt}
\setlength{\belowdisplayskip}{1pt}
\setlength{\belowdisplayshortskip}{0pt}
\setlength{\jot}{0pt}
\textbf{Out}: Discount ${\Db}_j$ for powered counts $\Cb^{n,(\rho_j)}$ and associated leftover weight $\gamma_j$
 \\[-0.4cm]
  \begin{algorithmic}[1]   
  \STATE Set $\Db_j(w_i, w^{i-1}_{i-n+1}) = d_\ast c(w_i, w^{i-1}_{i-n+1})^{\rho_{j+1}}$. 
  \STATE
  \begin{eqnarray}
  \gamma_j(w_i, w^{i-1}_{i-n+1})  = \frac{d_{\ast} {\sum_{w_i} c(w_i, w^{i-1}_{i-n+1})^{\rho_{j+1}}}}{{\sum_{w_i} c(w_i, w^{i-1}_{i-n+1})^{\rho_{j}}}} \notag
  \end{eqnarray} 
  \end{algorithmic}
  \label{alg:computed}   
\end{algorithm}

\begin{algorithm}[t!]
\caption{\small Compute $\Zb$}
\textbf{In}: Count tensor $\Cb^n$, power $\rho$, discounts $\Db$, rank $\kappa$ \\
\setlength{\abovedisplayskip}{2pt}
\setlength{\abovedisplayshortskip}{0pt}
\setlength{\belowdisplayskip}{1pt}
\setlength{\belowdisplayshortskip}{0pt}
\setlength{\jot}{0pt}
\textbf{Out}: Discounted low rank conditional probability table $\Zb_{\Db}^{(\rho,\kappa)}(w_i | w^{i-1}_{i-n+1})$ (represented implicitly) \\[-0.4cm]
  \begin{algorithmic}[1]    
    \STATE Compute powered counts $\Cb^{n,(\cdot \rho)}$.
      \STATE Compute denominators $\sum_{w_i} c(w_i, w^{i-1}_{i-n+1})^{\rho}$ $\forall w^{i-1}_{i-n+1}$ s.t. $c(w^{i-1}_{i-n+1}) > 0$. 
      \STATE Compute discounted powered counts $\Cb_{\Db}^{n,(\cdot \rho)} = \Cb^{n, (\cdot \rho)} - \Db$.
      \STATE \hspace{-2mm} For each slice  $\Mb_{\tilde{w}^{i-1}_{i-n+2}} := \Cb_{\Db}^{n, (\cdot \rho)}(:,  \tilde{w}^{i-1}_{i-n+2} ,:)$  compute 
      \begin{align*}
      \Mb^{(\kappa)}  := \min_{\Ab \geq 0: rank(\Ab) = \kappa} \| \Mb_{\tilde{w}^{i-1}_{i-n+2}}   - \Ab \|_{KL} \notag \\
      \textrm{(stored implicitly as $\Mb^{(\kappa)} = \Lb \Rb$)}
      \end{align*}
      \hspace{0.5cm} Set $\Zb_{\Db}^{(\rho,\kappa)}(:, \tilde{w}^{i-1}_{i-n+2}, :) = \Mb^{(\kappa)}$ 
      \STATE Note that  
      \begin{eqnarray*}
      \Zb_{\Db}^{(\rho,\kappa)}(w_i | w^{i-1}_{i-n+1}) =  \frac{\Zb_{\Db}^{(\rho,\kappa)}(w_i, w^{i-1}_{i-n+1})}{\sum_{w_i} c(w_i, w^{i-1}_{i-n+1})^{\rho}}
      \end{eqnarray*}
       \end{algorithmic}
  \label{alg:computez}   
\end{algorithm}

The next step is to compute low rank approximations of
$\Yb_{\Db_{j}}^{(\rho_j)}$  to obtain $\Zb_{\Db_j}$ such that
the intermediate marginal constraint in Eq.~\ref{eq:intermargconst} is
preserved. This constraint trivially holds for the
intermediate ensemble $P_{\textrm{pwr}}(w_i | w_{i-1})$ due to
how the discounts were derived in \S~\ref{sec:computedisc}.
%
For our running bigram example, define $\Zb_{\Db_j}^{(\rho_j, \kappa_j)}$ to be the best rank $\kappa_j$ approximation to $\Yb_{\Db_j}^{(\rho_j, \kappa_j)}$ according to $gKL$ and let
\begin{flalign*}
 \Zb_{\Db_j}^{\rho_j, \kappa_j}(w_i | w_{i-1}) = \frac{\Zb_{\Db_j}^{\rho_j, \kappa_j}(w_i, w_{i-1})}{\sum_{w_i}c(w_i, w_{i-1})^{\rho_j}}
\end{flalign*}
Note that $\Zb_{\Db_j}^{\rho_j, \kappa_j}(w_i | w_{i-1})$ is a valid (discounted) conditional probability since $gKL$ preserves row/column sums so the denominator remains unchanged under the low rank approximation. Then using the fact that $\Zb^{(0, 1)}(w_i | w_{i-1}) = P^\alt(w_i)$ (Lemma~\ref{lem:power}) we can
embellish Eq.~\ref{eq:overall_bigram} as
\begin{flalign*}
& P_{\plre}(w_i | w_{i-1}) = P_{\Db_0}(w_i | w_{i-1}) +  \notag \\
&\gamma_0(w_{i-1}) \bigg ( \Zb_{\Db_1}^{(\rho_1, \kappa_1)} (w_i | w_{i-1})  + \gamma_{1}(w_{i-1})  P_{\textrm{alt}}(w_i) \bigg )
\end{flalign*}

Leveraging the form of the discounts and row/column sum preserving property of $gKL$, we then have the following lemma (the proof is in the supplementary material):
\begin{lemma}
Let $P_{\textrm{plre}}(w_i | w_{i-1})$ indicate the PLRE smoothed conditional probability as computed by Eq.~\ref{eq:overall_bigram} and Algorithms~\ref{alg:computed} and~\ref{alg:computez}. Then, the marginal constraint in Eq.~\ref{eq:intermargconst} holds.
\label{lem:marglemma}
\end{lemma}

\subsection{More general algorithm}

In general, the principles outlined in the previous sections hold for higher order $n$-grams. 
Assume that the discounts are computed according to Algorithm~\ref{alg:computed} with parameter $d_{\ast}$ and $\Zb_{\Db_j}^{(\rho_j, \kappa_j)}$ is computed according to Algorithm~\ref{alg:computez}. Note that, as shown in Algorithm~\ref{alg:computez}, for higher order $n$-grams, the $\Zb_{\Db_j}^{(\rho_j, \kappa_j)}$ are created by taking low rank approximations of slices of the (powered) count tensors (see Lemma~\ref{lem:power} for intuition).  
Eq.~\ref{eq:overall} can now be embellished:
\begin{flalign}
& P_{\textrm{plre}}(w_i | w^{i-1}_{i-n+1}) = P_{\Db_0}^{\textrm{alt}}(w_i | w^{i-1}_{i-n+1}) \notag \\
&+ \gamma_0(w^{i-1}_{i-n+1}) \bigg ( \Zb_{\Db_1}^{(\rho_1, \kappa_1)} (w_i | w^{i-1}_{i-n+1})  + .....  \notag \\
& + \gamma_{\eta-1}(w^{i-1}_{i-n+1}) \bigg  (  \Zb_{\Db_\eta}^{(\rho_\eta, \kappa_\eta)}  (w_i | w^{i-1}_{i-n+1})  \notag \\
&+ \gamma_{\eta}(w^{i-1}_{i-n+1}) \bigg (  P_{\textrm{plre}}(w_i | w^{i-1}_{i-n+2}) \bigg  ) \bigg ) ...  \bigg )
\label{eq:overallagain}
\end{flalign}
Lemma~\ref{lem:marglemma} also applies in this case and is given in Theorem 1 in the supplementary material.

\subsection{Links with KN Smoothing}
\label{sec:knlinks}
In this section, we explicitly show the relationship between PLRE and
KN smoothing.  Rewriting Eq.~\ref{eq:overallagain} in the following form:
\begin{eqnarray}
P_{\textrm{plre}}(w_i | w^{i-1}_{i-n+1}) = P_{\textrm{plre}}^{\textrm{terms}}(w_i | w^{i-1}_{i-n+1}) \notag \\
+ \gamma_{0:\eta}(w^{i-1}_{i-n+1}) P_{\textrm{plre}}(w_i | w^{i-1}_{i-n+2})
\label{eq:kncompare}
\end{eqnarray}
where $P^{\textrm{terms}}_{\textrm{plre}}(w_i | w^{i-1}_{i-n+1})$ contains the
terms in Eq.~\ref{eq:overallagain} except the last, and $\gamma_{0:\eta}(w^{i-1}_{i-n+1})   =  \prod_{h=0}^\eta \gamma_h(w^{i-1}_{i-n+1})$,
we can leverage the form of the discount, and using the fact that $\rho_{\eta+1} = 0$\footnote{for derivation see proof of Lemma~\ref{lem:marglemma} in the supplementary material}:
\begin{eqnarray}
\gamma_{0:\eta}(w^{i-1}_{i-n-1}) = \frac{{d_{\ast}}^{\eta+1}
  N_{+}(w^{i-1}_{i-n+1})}{c(w^{i-1}_{i-n+1})} \notag
\end{eqnarray}
With this form of $\gamma(\cdot)$, Eq.~\ref{eq:kncompare} is remarkably
similar to KN smoothing (Eq.~\ref{eq:kney}) if KN's discount parameter $D$ is chosen to equal ${(d_{\ast})}^{\eta+1}$.

The difference is that $P^{\textrm{alt}}(\cdot)$ has been replaced
with the alternate estimate $P_{\textrm{plre}}^{\textrm{terms}}(w_i |
w^{i-1}_{i-n+1})$, which have been enriched via the low rank
structure.  Since these alternate estimates were constructed via our
ensemble strategy they contain both very fine-grained dependencies
(the original $n$-grams) as well as coarser dependencies (the lower rank $n$-grams) and is thus fundamentally different than simply taking a single matrix/tensor decomposition of the trigram/bigram matrices.

Moreover, it provides a natural way of setting $d_{\ast}$ based on
the Good-Turing (GT) estimates employed by KN smoothing. In particular, we
can set $d_{\ast}$ to be the ${(\eta+1)}^{\textrm{th}}$ root of the KN
discount $D$ that can be estimated via the GT estimates.

\subsection{Computational Considerations}
PLRE scales well even as the order $n$ increases. To compute a low rank bigram, one low rank approximation of a $V \times V$ matrix is required. For the low rank trigram, we need to compute a low rank approximation
of each slice $\Cb_{\Db}^{n,(\cdot p)}(:, \tilde{w}_{i-1},:) \,\, \forall \tilde{w}_{i-1}$. While this may seem daunting at first, in practice the \textit{size} of each slice (number of non-zero rows/columns) is usually much, much smaller than $V$,
keeping the computation tractable.

Similarly, PLRE also evaluates conditional probabilities at evaluation
time efficiently. As shown in Algorithm~\ref{alg:computez}, the
normalizer can be precomputed on the sparse powered matrix/tensor. As
a result our test complexity is $\mathcal{O}(\sum_{i=1}^{\eta_{\textrm{total}}}
\kappa_i)$ where $\eta_{\textrm{total}}$ is the total number of matrices/tensors in the
ensemble. While this is larger  than Kneser Ney's practically constant
complexity of $\mathcal{O}(n)$, it is much faster than other recent
methods for language modeling such as neural networks and conditional exponential
family models where exact computation of the normalizing constant costs $O(V)$.  
\vspace{-2mm}
\section{Experiments}
\label{sec:experiments}
To evaluate PLRE, we compared  its performance
on English and Russian corpora with several variants of KN smoothing, class-based models, and the log-bilinear neural language model~\cite{mnih2007three}.  We evaluated with perplexity in most of our experiments, but also provide results evaluated with BLEU \cite{Papineni2002} on a downstream machine translation (MT) task. 
We have made the code for our approach publicly available~\footnote{http://www.cs.cmu.edu/$\sim$apparikh/plre.html\label{website}}.

\begin{table*}
\small
\begin{center}
\begin{tabular}{|c|c|c|c|c|c|c|}
	\hline
Dataset &  class-1024(3) & BO-KN(3) & int-KN(3) & BO-MKN(3) & int-MKN(3) & PLRE(3)   \\
	\hline
Small-English Dev & 115.64  &  99.20 & 99.73 & 99.95 & 95.63 & \textbf{91.18}   \\
Small-English Test & 119.70  & 103.86 &   104.56   &  104.55  & 100.07  & \textbf{95.15}  \\
Small-Russian Dev & 286.38  & 281.29 & 265.71 & 287.19 & 263.25 & \textbf{241.66}  \\
Small-Russian Test & 284.09 & 277.74 & 262.02 & 283.70 & 260.19  & \textbf{238.96}  \\
	\hline
\end{tabular}
\end{center}
\vspace{-2mm}
\caption{Perplexity results on small corpora for all methods.}
\label{fig:table-small}
\vspace{-2mm}
\end{table*}


%

To build the hard class-based LMs, we utilized \texttt{mkcls}\footnote{http://code.google.com/p/giza-pp/}, a tool to train word classes that uses the maximum likelihood criterion \cite{Och1995} for classing. 
We subsequently trained trigram class language models on these classes (corresponding to $2^{\textrm{nd}}$-order HMMs) using SRILM \cite{Stolcke2002}, with KN-smoothing for the class transition probabilities. 
SRILM was also used for the baseline KN-smoothed models.  

For our MT evaluation, we built a hierarchical phrase translation \cite{Chiang2007} system using \texttt{cdec} \cite{dyer2010cdec}.  
The KN-smoothed models in the MT experiments were compiled using KenLM \cite{Heafield2011}. 


\vspace{-1mm}
\subsection{Datasets}
\label{sec:datasets}
\vspace{-1mm}

For the perplexity experiments, we evaluated our proposed approach on 4 datasets, 2 in English and 2 in Russian.  
In all cases, the singletons were replaced with ``$<$unk$>$'' tokens in the training corpus, and any word not in the vocabulary was replaced with this token during evaluation. 
There is a general dearth of evaluation on large-scale corpora in morphologically rich languages such as Russian, and thus we have made the processed Large-Russian corpus available for comparison~\footref{website}. 

\begin{itemizesquish}{-0.3em}{0.5em}
\item \textbf{Small-English}: APNews corpus \cite{Bengio2003}: Train - 14 million words, Dev - 963,000, Test - 963,000. Vocabulary- 18,000 types. 
\item \textbf{Small-Russian}: Subset of Russian news commentary data from 2013 WMT translation
task\footnote{http://www.statmt.org/wmt13/training-monolingual-nc-v8.tgz}: Train- 3.5 million words, Dev - 400,000 Test - 400,000. Vocabulary - 77,000 types. 
\item \textbf{Large-English}: English Gigaword, Training - 837 million words, Dev - 8.7 million,  Test - 8.7 million. Vocabulary- 836,980 types. 
\item \textbf{Large-Russian}: Monolingual data from WMT 2013 task. Training - 521 million words, Validation - 50,000, Test - 50,000. Vocabulary- 1.3 million types.
\end{itemizesquish}

For the MT evaluation, we used the parallel data from the WMT 2013 shared task, excluding the Common Crawl corpus data. The newstest2012 and newstest2013 evaluation sets were used as the development and test sets respectively. 


\vspace{-1mm}
\subsection{Small Corpora}
\vspace{-1mm}

For the class-based baseline LMs, the number of classes was
selected from $\{32, 64, 128, 256, 512, 1024\}$ (Small-English) and $\{512, 1024\}$ (Small-Russian).  We could not go higher due to the computationally laborious process of hard clustering.  For Kneser-Ney, we explore four different variants: back-off (BO-KN) interpolated (int-KN), modified back-off (BO-MKN), and modified interpolated (int-MKN). Good-Turing estimates were used for discounts.  All models trained on the small corpora are of order 3 (trigrams).  

For PLRE, we used one low rank bigram and one low rank trigram in addition to the MLE $n$-gram estimates. The powers of the intermediate matrices/tensors were fixed to be $0.5$ and the discounts were set to be square roots of the Good Turing estimates (as explained in \S~\ref{sec:knlinks}). The ranks were tuned on the development set. For Small-English, the ranges were $\{1\textrm{e}-3, 5\textrm{e}-3\}$ (as a fraction of the vocabulary size) for both the low rank bigram and low rank trigram models.  For Small-Russian the ranges were $\{5\textrm{e}-4, 1\textrm{e}-3\}$ for both the low rank bigram and the low rank trigram models. 

The results are shown in Table~\ref{fig:table-small}. The best class-based LM is reported, but is not competitive with the KN baselines. PLRE outperforms all of the baselines comfortably. Moreover, PLRE's performance over the baselines is highlighted in Russian. With larger vocabulary sizes, the low rank approach is more 
effective as it can capture linguistic similarities between rare and common words.  

\begin{figure}[!tpb]
 \centerline{\includegraphics[width=8cm]{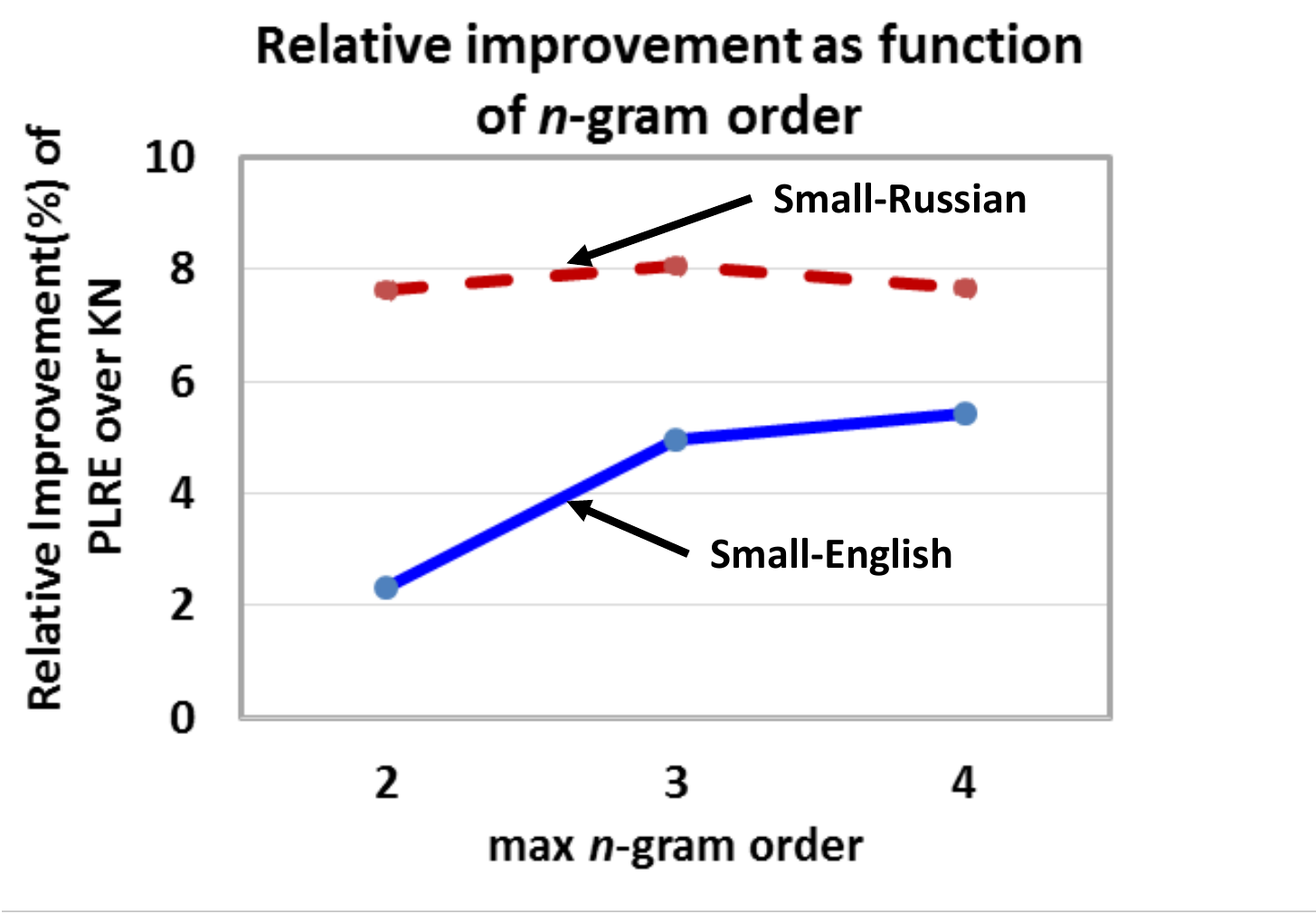}}
\caption{ Relative percentage improvement of PLRE over int-MKN as the maximum $n$-gram order for both methods is increased.}
\label{fig:ngram-order}
\vspace{-3mm}
 \end{figure}

Next we discuss how the maximum $n$-gram order affects performance. Figure~\ref{fig:ngram-order} shows the relative percentage improvement of our approach over int-MKN as the order is increased from 2 to 4 for both methods. The Small-English dataset has a rather small vocabulary compared to the  number of tokens, leading to lower data sparsity in the bigram. Thus the PLRE improvement is small for $\textrm{order}=2$, but more substantial for $\textrm{order}=3$.
On the other hand, for the Small-Russian dataset, the vocabulary size is much larger and consequently the bigram counts are sparser. This leads to similar improvements for all orders (which are larger than that for Small-English).

On both these datasets, we also experimented with tuning the discounts for int-MKN to see if the baseline could be improved with more careful choices of discounts.
However, this achieved only marginal gains (reducing the perplexity to $98.94$ on the Small-English test set and $259.0$ on the Small-Russian test set).

\textbf{Comparison to LBL~\cite{mnih2007three}}:  \newcite{mnih2007three} evaluate on the Small-English 
dataset (but remove end markers and concatenate the sentences). They obtain perplexities $117.0$ and $107.8$ using contexts of size 5 and 10 respectively. With this preprocessing, a 4-gram (context 3) PLRE achieves $108.4$ perplexity.

\vspace{-1mm}
\subsection{Large Corpora}
\vspace{-1mm}
Results on the larger corpora for the top 2 performing methods ``PLRE" and ``int-MKN" are presented in Table~\ref{tab:large-results}.  
Due to the larger training size, we use 4-gram models in these experiments. 
However, including the low rank 4-gram tensor provided little gain and therefore, the 4-gram PLRE only has additional low rank bigram and low rank trigram matrices/tensors. As above, ranks were tuned on the development set. For Large-English, the ranges were $\{1\textrm{e}-4, 5\textrm{e}-4, 1\textrm{e}-3\}$ (as a fraction of the vocabulary size) for both the low rank bigram and low rank trigram models.  For Small-Russian the ranges were $\{1\textrm{e}-5, 5\textrm{e}-5, 1\textrm{e}-4\}$ for both the low rank bigram and the low rank trigram models. 
For statistical validity, 10 test sets of size equal to the original test set were generated by randomly sampling sentences with replacement from the original test set. Our method outperforms ``int-MKN" with gains similar to that on the smaller datasets. As shown in Table~\ref{tab:timing}, our method obtains fast training times even for large datasets.

{\footnotesize
\begin{table}
\small
\begin{center}
\begin{tabular}{|c|c|c|}
	\hline
Dataset &  int-MKN(4) & PLRE(4)   \\
	\hline
Large-English Dev & 73.21 & \textbf{71.21}   \\
Large-English Test & 77.90 $\pm$  0.203  &  \textbf{75.66} $\pm$ 0.189  \\
Large-Russian Dev & 326.9  & \textbf{297.11}  \\
Large-Russian Test & 289.63 $\pm$ 6.82  & \textbf{264.59} $\pm$ 5.839   \\
	\hline
\end{tabular}
\end{center}
\vspace{-2mm}
\caption{Mean perplexity results on large corpora, with standard deviation.}
\label{tab:large-results}
\vspace{-2mm}
\end{table}
}

\vspace{-1mm}
\section{Machine Translation Task}
\vspace{-1mm}
\label{sec:mt}

\begin{table}
\small
\begin{center}
\begin{tabular}{|c|c|}
\hline
Dataset &  PLRE Training Time   \\
\hline
Small-English  & 3.96 min ( order 3) / 8.3 min (order 4)  \\
Small-Russian  & 4.0 min (order 3) / 4.75 min (order 4)  \\
Large-English  & 3.2 hrs (order 4)  \\
Large-Russian  & 8.3 hrs (order 4) \\
\hline
\end{tabular}
\end{center}
\vspace{-2mm}
\caption{PLRE training times for a fixed parameter setting\footnotemark. 8 Intel Xeon CPUs  were used.} 
\label{tab:timing}
\vspace{-2mm}
\end{table}
\begin{table}[tbp!]
\small
\begin{center}
\begin{tabular}{|c|c|}
\hline
Method &  BLEU   \\
\hline
int-MKN(4) & 17.63 $\pm$  0.11 \\
PLRE(4) & 17.79 $\pm$ 0.07  \\
Smallest Diff & PLRE+0.05  \\
Largest Diff & PLRE+0.29  \\
\hline
\end{tabular}
\end{center}
\vspace{-2mm}
\caption{Results on English-Russian translation task (mean $\pm$ stdev). See text for details.}
\label{fig:MT-results}
\vspace{-2mm}
\end{table}
\footnotetext{As described earlier, only the ranks need to be tuned, so only 2-3 low rank bigrams and 2-3 low rank trigrams need to be computed (and combined depending on the setting).}

Table~\ref{fig:MT-results} presents results for the MT task, translating from English to Russian\footnote{the best score at WMT 2013 was 19.9 \cite{Bojar2013}}.
We used MIRA~\cite{chiang2008online} to learn the feature weights. 
To control for the randomness in MIRA, we avoid retuning when switching LMs - the set of feature weights obtained using int-MKN is the same, only the language model changes. 
The procedure is repeated 10 times to control for optimizer instability \cite{Clark2011}.  
Unlike other recent approaches where an additional feature weight is tuned for the proposed model and used in conjunction with KN smoothing \cite{Vaswani2013}, our aim is to show the improvements that PLRE provides as a substitute for KN.  
On average, PLRE outperforms the KN baseline by 0.16 BLEU, and this improvement is consistent in that PLRE never gets a worse BLEU score.  
\vspace{-2mm}
\section{Related Work}
\vspace{-2mm}
\label{sec:related}
Recent attempts to revisit the language modeling problem have largely come from two directions: Bayesian nonparametrics and neural networks. \newcite{teh2006hierarchical} and \newcite{goldwater06} discovered the connection between interpolated Kneser Ney and the hierarchical Pitman-Yor process. These have led to generalizations that account for domain effects \cite{Wood2009} and unbounded contexts \cite{wood2009stochastic}.

The idea of using neural networks for language modeling is not new~\cite{Miikkulainen1991}, but recent efforts~\cite{mnih2007three,Mikolov2010} have achieved impressive performance.  These methods can be quite expensive to train and query (especially as the vocabulary size increases).  Techniques such as noise contrastive estimation \cite{Gutmann2012,Mnih2012,Vaswani2013}, subsampling \cite{Xu2011}, or careful engineering approaches for maximum entropy LMs (which can also be applied to neural networks) \cite{Wu2000} have improved training of these models, but querying the probability of the next word given still requires explicitly normalizing over the vocabulary, which is expensive for big corpora or in languages with a large number of word types. \newcite{Mnih2012} and \newcite{Vaswani2013} propose setting the normalization constant to 1, but this is approximate  and thus can only be used for downstream evaluation, not for perplexity computation.  An alternate technique is to use word-classing~\cite{goodman2001classes,mikolov2011extensions}, which can reduce the cost of exact normalization to $O(\sqrt{V})$.
In contrast, our approach is much more scalable, since it is trivially parallelized in training and does not require explicit normalization during evaluation.


There are a few low rank approaches~\cite{saul1997aggregate,bellegarda2000,hutchinson2011low}, but they are only effective in restricted settings (e.g. small training sets, or corpora divided into documents) and do not generally perform comparably to state-of-the-art models. 
\newcite{Roark2013} also use the idea of marginal constraints for re-estimating back-off parameters for heavily-pruned language models, whereas we use this concept to estimate $n$-gram specific discounts.   
\vspace{-2mm}
\section{Conclusion}
\vspace{-2mm}
We presented power low rank ensembles, a technique that
generalizes existing $n$-gram smoothing techniques to
non-integer $n$.  By using ensembles of sparse as well as low rank matrices and tensors, our method
captures both the fine-grained and coarse structures in word sequences.
Our discounting strategy preserves the marginal constraint and thus generalizes Kneser Ney, and under
slight changes can also extend other smoothing methods such as deleted-interpolation/Jelinek-Mercer smoothing.
Experimentally, PLRE convincingly outperforms Kneser-Ney smoothing as well as class-based baselines.

\section*{Acknowledgements}

This work was supported by NSF IIS1218282, NSF IIS1218749, NSF
IIS1111142, NIH R01GM093156, the U.~S.~Army Research Laboratory
and the U.~S.~Army Research Office under contract/grant number
W911NF-10-1-0533, the NSF
Graduate Research Fellowship Program under
Grant No. 0946825 (NSF Fellowship to APP), and a grant from Ebay Inc. (to AS).


\vspace{-2mm}
\bibliographystyle{acl}
\bibliography{thesis-bib,tacl2013}

\begin{thebibliography}{}

\bibitem[\protect\citename{Bellegarda}2000]{bellegarda2000}
Jerome~R. Bellegarda.
\newblock 2000.
\newblock Large vocabulary speech recognition with multispan statistical
  language models.
\newblock {\em IEEE Transactions on Speech and Audio Processing}, 8(1):76--84.

\bibitem[\protect\citename{Bengio \bgroup et al.\egroup }2003]{Bengio2003}
Yoshua Bengio, R{\'e}jean Ducharme, Pascal Vincent, and Christian Janvin.
\newblock 2003.
\newblock A neural probabilistic language model.
\newblock {\em J. Mach. Learn. Res.}, 3:1137--1155, March.

\bibitem[\protect\citename{Bojar \bgroup et al.\egroup }2013]{Bojar2013}
Ond\v{r}ej Bojar, Christian Buck, Chris Callison-Burch, Christian Federmann,
  Barry Haddow, Philipp Koehn, Christof Monz, Matt Post, Radu Soricut, and
  Lucia Specia.
\newblock 2013.
\newblock Findings of the 2013 {Workshop on Statistical Machine Translation}.
\newblock In {\em Proceedings of the Eighth Workshop on Statistical Machine
  Translation}, pages 1--44, Sofia, Bulgaria, August. Association for
  Computational Linguistics.

\bibitem[\protect\citename{Cai \bgroup et al.\egroup }2010]{cai2010singular}
Jian-Feng Cai, Emmanuel~J Cand{\`e}s, and Zuowei Shen.
\newblock 2010.
\newblock A singular value thresholding algorithm for matrix completion.
\newblock {\em SIAM Journal on Optimization}, 20(4):1956--1982.

\bibitem[\protect\citename{Cand{\`e}s and Recht}2009]{candes2009exact}
Emmanuel~J Cand{\`e}s and Benjamin Recht.
\newblock 2009.
\newblock Exact matrix completion via convex optimization.
\newblock {\em Foundations of Computational mathematics}, 9(6):717--772.

\bibitem[\protect\citename{Chen and Goodman}1999]{Chen1999}
Stanley~F. Chen and Joshua Goodman.
\newblock 1999.
\newblock An empirical study of smoothing techniques for language modeling.
\newblock {\em Computer Speech {\&} Language}, 13(4):359--393.

\bibitem[\protect\citename{Chen and Rosenfeld}2000]{chen2000survey}
Stanley~F Chen and Ronald Rosenfeld.
\newblock 2000.
\newblock A survey of smoothing techniques for me models.
\newblock {\em Speech and Audio Processing, IEEE Transactions on}, 8(1):37--50.

\bibitem[\protect\citename{Chen}2009]{Chen2009}
Stanley~F. Chen.
\newblock 2009.
\newblock Shrinking exponential language models.
\newblock In {\em Proceedings of Human Language Technologies: The 2009 Annual
  Conference of the North American Chapter of the Association for Computational
  Linguistics}, NAACL '09, pages 468--476, Stroudsburg, PA, USA. Association
  for Computational Linguistics.

\bibitem[\protect\citename{Chiang \bgroup et al.\egroup
  }2008]{chiang2008online}
David Chiang, Yuval Marton, and Philip Resnik.
\newblock 2008.
\newblock Online large-margin training of syntactic and structural translation
  features.
\newblock In {\em Proceedings of the Conference on Empirical Methods in Natural
  Language Processing}, pages 224--233. Association for Computational
  Linguistics.

\bibitem[\protect\citename{Chiang}2007]{Chiang2007}
David Chiang.
\newblock 2007.
\newblock Hierarchical phrase-based translation.
\newblock {\em Comput. Linguist.}, 33(2):201--228, June.

\bibitem[\protect\citename{Clark \bgroup et al.\egroup }2011]{Clark2011}
Jonathan~H. Clark, Chris Dyer, Alon Lavie, and Noah~A. Smith.
\newblock 2011.
\newblock Better hypothesis testing for statistical machine translation:
  Controlling for optimizer instability.
\newblock In {\em Proceedings of the 49th Annual Meeting of the Association for
  Computational Linguistics: Human Language Technologies: Short Papers - Volume
  2}, HLT '11, pages 176--181.

\bibitem[\protect\citename{Dyer \bgroup et al.\egroup }2010]{dyer2010cdec}
Chris Dyer, Jonathan Weese, Hendra Setiawan, Adam Lopez, Ferhan Ture, Vladimir
  Eidelman, Juri Ganitkevitch, Phil Blunsom, and Philip Resnik.
\newblock 2010.
\newblock cdec: A decoder, alignment, and learning framework for finite-state
  and context-free translation models.
\newblock In {\em Proceedings of the ACL 2010 System Demonstrations}, pages
  7--12. Association for Computational Linguistics.

\bibitem[\protect\citename{Goldwater \bgroup et al.\egroup }2006]{goldwater06}
Sharon Goldwater, Thomas Griffiths, and Mark Johnson.
\newblock 2006.
\newblock Interpolating between types and tokens by estimating power-law
  generators.
\newblock In {\em Advances in Neural Information Processing Systems},
  volume~18.

\bibitem[\protect\citename{Goodman}2001]{goodman2001classes}
Joshua Goodman.
\newblock 2001.
\newblock Classes for fast maximum entropy training.
\newblock In {\em Acoustics, Speech, and Signal Processing, 2001.
  Proceedings.(ICASSP'01). 2001 IEEE International Conference on}, volume~1,
  pages 561--564. IEEE.

\bibitem[\protect\citename{Gutmann and Hyv\"arinen}2012]{Gutmann2012}
Michael Gutmann and Aapo Hyv\"arinen.
\newblock 2012.
\newblock Noise-contrastive estimation of unnormalized statistical models, with
  applications to natural image statistics.
\newblock {\em Journal of Machine Learning Research}, 13:307--361.

\bibitem[\protect\citename{Heafield}2011]{Heafield2011}
Kenneth Heafield.
\newblock 2011.
\newblock {KenLM:} faster and smaller language model queries.
\newblock In {\em Proceedings of the {EMNLP} 2011 Sixth Workshop on Statistical
  Machine Translation}, pages 187--197, Edinburgh, Scotland, United Kingdom,
  July.

\bibitem[\protect\citename{Ho and Van~Dooren}2008]{ho2008non}
Ngoc-Diep Ho and Paul Van~Dooren.
\newblock 2008.
\newblock Non-negative matrix factorization with fixed row and column sums.
\newblock {\em Linear Algebra and its Applications}, 429(5):1020--1025.

\bibitem[\protect\citename{Hutchinson \bgroup et al.\egroup
  }2011]{hutchinson2011low}
Brian Hutchinson, Mari Ostendorf, and Maryam Fazel.
\newblock 2011.
\newblock Low rank language models for small training sets.
\newblock {\em Signal Processing Letters, IEEE}, 18(9):489--492.

\bibitem[\protect\citename{Jelinek and Mercer}1980]{jelinek1980interpolated}
Frederick Jelinek and Robert Mercer.
\newblock 1980.
\newblock Interpolated estimation of markov source parameters from sparse data.
\newblock {\em Pattern recognition in practice}.

\bibitem[\protect\citename{Kneser and Ney}1995]{kneser1995improved}
Reinhard Kneser and Hermann Ney.
\newblock 1995.
\newblock Improved backing-off for $m$-gram language modeling.
\newblock In {\em Acoustics, Speech, and Signal Processing, 1995. ICASSP-95.,
  1995 International Conference on}, volume~1, pages 181--184. IEEE.

\bibitem[\protect\citename{Koehn}2010]{koehn2010}
Philipp Koehn.
\newblock 2010.
\newblock {\em Statistical Machine Translation}.
\newblock Cambridge University Press, New York, NY, USA, 1st edition.

\bibitem[\protect\citename{Koren \bgroup et al.\egroup }2009]{koren2009matrix}
Yehuda Koren, Robert Bell, and Chris Volinsky.
\newblock 2009.
\newblock Matrix factorization techniques for recommender systems.
\newblock {\em Computer}, 42(8):30--37.

\bibitem[\protect\citename{Lee and Seung}2001]{seung2001algorithms}
Daniel~D. Lee and H.~Sebastian Seung.
\newblock 2001.
\newblock Algorithms for non-negative matrix factorization.
\newblock {\em Advances in Neural Information Processing Systems}, 13:556--562.

\bibitem[\protect\citename{Mackey \bgroup et al.\egroup
  }2011]{mackey2011divide}
Lester Mackey, Ameet Talwalkar, and Michael~I Jordan.
\newblock 2011.
\newblock Divide-and-conquer matrix factorization.
\newblock {\em arXiv preprint arXiv:1107.0789}.

\bibitem[\protect\citename{Manning and
  Sch{\"u}tze}1999]{manning1999foundations}
Christopher~D Manning and Hinrich Sch{\"u}tze.
\newblock 1999.
\newblock {\em Foundations of statistical natural language processing}, volume
  999.
\newblock MIT Press.

\bibitem[\protect\citename{Miikkulainen and Dyer}1991]{Miikkulainen1991}
Risto Miikkulainen and Michael~G. Dyer.
\newblock 1991.
\newblock Natural language processing with modular pdp networks and distributed
  lexicon.
\newblock {\em Cognitive Science}, 15:343--399.

\bibitem[\protect\citename{Mikolov \bgroup et al.\egroup }2010]{Mikolov2010}
Tomáš Mikolov, Martin Karafiát, Lukáš Burget, Jan Černocký, and Sanjeev
  Khudanpur.
\newblock 2010.
\newblock Recurrent neural network based language model.
\newblock In {\em Proceedings of the 11th Annual Conference of the
  International Speech Communication Association (INTERSPEECH 2010)}, volume
  2010, pages 1045--1048. International Speech Communication Association.

\bibitem[\protect\citename{Mikolov \bgroup et al.\egroup
  }2011]{mikolov2011extensions}
Tomas Mikolov, Stefan Kombrink, Lukas Burget, JH~Cernocky, and Sanjeev
  Khudanpur.
\newblock 2011.
\newblock Extensions of recurrent neural network language model.
\newblock In {\em Acoustics, Speech and Signal Processing (ICASSP), 2011 IEEE
  International Conference on}, pages 5528--5531. IEEE.

\bibitem[\protect\citename{Mnih and Hinton}2007]{mnih2007three}
Andriy Mnih and Geoffrey Hinton.
\newblock 2007.
\newblock Three new graphical models for statistical language modelling.
\newblock In {\em Proceedings of the 24th international conference on Machine
  learning}, pages 641--648. ACM.

\bibitem[\protect\citename{Mnih and Teh}2012]{Mnih2012}
A.~Mnih and Y.~W. Teh.
\newblock 2012.
\newblock A fast and simple algorithm for training neural probabilistic
  language models.
\newblock In {\em Proceedings of the International Conference on Machine
  Learning}.

\bibitem[\protect\citename{Nelakanti \bgroup et al.\egroup
  }2013]{Nelakanti2013}
Anil~Kumar Nelakanti, Cedric Archambeau, Julien Mairal, Francis Bach, and
  Guillaume Bouchard.
\newblock 2013.
\newblock Structured penalties for log-linear language models.
\newblock In {\em Proceedings of the 2013 Conference on Empirical Methods in
  Natural Language Processing}, pages 233--243, Seattle, Washington, USA,
  October. Association for Computational Linguistics.

\bibitem[\protect\citename{Ney \bgroup et al.\egroup }1994]{Ney1994}
Hermann Ney, Ute Essen, and Reinhard Kneser.
\newblock 1994.
\newblock {On Structuring Probabilistic Dependencies in Stochastic Language
  Modelling}.
\newblock {\em Computer Speech and Language}, 8:1--38.

\bibitem[\protect\citename{Och}1995]{Och1995}
Franz~Josef Och.
\newblock 1995.
\newblock Maximum-likelihood-sch\"atzung von wortkategorien mit verfahren der
  kombinatorischen optimierung.
\newblock Bachelor's thesis ({Studienarbeit}), University of Erlangen.

\bibitem[\protect\citename{Papineni \bgroup et al.\egroup }2002]{Papineni2002}
Kishore Papineni, Salim Roukos, Todd Ward, and Wei jing Zhu.
\newblock 2002.
\newblock Bleu: a method for automatic evaluation of machine translation.
\newblock pages 311--318.

\bibitem[\protect\citename{Rabiner and Juang}1993]{rabiner1993fundamentals}
Lawrence Rabiner and Biing-Hwang Juang.
\newblock 1993.
\newblock Fundamentals of speech recognition.

\bibitem[\protect\citename{Roark \bgroup et al.\egroup }2013]{Roark2013}
Brian Roark, Cyril Allauzen, and Michael Riley.
\newblock 2013.
\newblock Smoothed marginal distribution constraints for language modeling.
\newblock In {\em Proceedings of the 51st Annual Meeting of the Association for
  Computational Linguistics (ACL)}, pages 43--52.

\bibitem[\protect\citename{Salakhutdinov and
  Mnih}2008]{salakhutdinov2008bayesian}
Ruslan Salakhutdinov and Andriy Mnih.
\newblock 2008.
\newblock Bayesian probabilistic matrix factorization using {Markov} chain
  {Monte Carlo}.
\newblock In {\em Proceedings of the 25th international conference on Machine
  learning}, pages 880--887. ACM.

\bibitem[\protect\citename{Saul and Pereira}1997]{saul1997aggregate}
Lawrence Saul and Fernando Pereira.
\newblock 1997.
\newblock Aggregate and mixed-order markov models for statistical language
  processing.
\newblock In {\em Proceedings of the second conference on empirical methods in
  natural language processing}, pages 81--89. Somerset, New Jersey: Association
  for Computational Linguistics.

\bibitem[\protect\citename{Stolcke}2002]{Stolcke2002}
Andreas Stolcke.
\newblock 2002.
\newblock {SRILM - An Extensible Language Modeling Toolkit}.
\newblock In {\em Proceedings of the International Conference in Spoken
  Language Processing}.

\bibitem[\protect\citename{Su and Khoshgoftaar}2009]{su2009survey}
Xiaoyuan Su and Taghi~M Khoshgoftaar.
\newblock 2009.
\newblock A survey of collaborative filtering techniques.
\newblock {\em Advances in artificial intelligence}, 2009:4.

\bibitem[\protect\citename{Teh}2006]{teh2006hierarchical}
Yee~Whye Teh.
\newblock 2006.
\newblock A hierarchical bayesian language model based on pitman-yor processes.
\newblock In {\em Proceedings of the 21st International Conference on
  Computational Linguistics and the 44th annual meeting of the Association for
  Computational Linguistics}, pages 985--992. Association for Computational
  Linguistics.

\bibitem[\protect\citename{Vaswani \bgroup et al.\egroup }2013]{Vaswani2013}
Ashish Vaswani, Yinggong Zhao, Victoria Fossum, and David Chiang.
\newblock 2013.
\newblock Decoding with large-scale neural language models improves
  translation.
\newblock In {\em Proceedings of the 2013 Conference on Empirical Methods in
  Natural Language Processing}, pages 1387--1392, Seattle, Washington, USA,
  October. Association for Computational Linguistics.

\bibitem[\protect\citename{Wood and Teh}2009]{Wood2009}
F.~Wood and Y.W. Teh.
\newblock 2009.
\newblock A hierarchical nonparametric {B}ayesian approach to statistical
  language model domain adaptation.
\newblock In {\em Artificial Intelligence and Statistics}, pages 607--614.

\bibitem[\protect\citename{Wood \bgroup et al.\egroup
  }2009]{wood2009stochastic}
Frank Wood, C{\'e}dric Archambeau, Jan Gasthaus, Lancelot James, and Yee~Whye
  Teh.
\newblock 2009.
\newblock A stochastic memoizer for sequence data.
\newblock In {\em Proceedings of the 26th Annual International Conference on
  Machine Learning}, pages 1129--1136. ACM.

\bibitem[\protect\citename{Wu and Khudanpur}2000]{Wu2000}
Jun Wu and Sanjeev Khudanpur.
\newblock 2000.
\newblock Efficient training methods for maximum entropy language modeling.
\newblock In {\em Interspeech}, pages 114--118.

\bibitem[\protect\citename{Xu \bgroup et al.\egroup }2011]{Xu2011}
Puyang Xu, Asela Gunawardana, and Sanjeev Khudanpur.
\newblock 2011.
\newblock Efficient subsampling for training complex language models.
\newblock In {\em Proceedings of the Conference on Empirical Methods in Natural
  Language Processing}, EMNLP '11, pages 1128--1136, Stroudsburg, PA, USA.
  Association for Computational Linguistics.

\bibitem[\protect\citename{Zipf}1949]{Zipf1949}
George Zipf.
\newblock 1949.
\newblock Human behaviour and the principle of least-effort.
\newblock Addison-Wesley, Cambridge, MA.

\end{thebibliography}


\begin{thebibliography}{}

\bibitem[\protect\citename{Chen and Goodman}1999]{Chen1999}
Stanley~F. Chen and Joshua Goodman.
\newblock 1999.
\newblock An empirical study of smoothing techniques for language modeling.
\newblock {\em Computer Speech {\&} Language}, 13(4):359--393.

\end{thebibliography}
\end{document}


\maketitle

The primary purpose of the supplementary material is to provide a proof of Lemma 4. We also show that Lemma 4 extends to $n > 2$.

%

\section{Proof of Lemma 4}

\begin{lemma}
Let $P_{\textrm{plre}}(w_i | w_{i-1})$ indicate the PLRE smoothed conditional probability and $\Phat(w)$ indicate the MLE probability of $w$. Then,
\begin{flalign}
\Phat(w) = \sum_{w_{i-1}} P_{\textrm{plre}}(w_i | w_{i-1}) \Phat(w_{i-1})
\end{flalign}

\begin{proof}

Assume the following more general form where multiple low rank matrices can be used i.e.:

\begin{flalign}
& P_{\textrm{plre}}(w_i | w_{i-1}) = P_{\Db_0}^{\textrm{alt}}(w_i | w_{i-1}) + \gamma_0(w_{i-1}) \bigg ( \Zb_{\Db_1}^{(\rho_1, \kappa_1)} (w_i | w_{i-1})  + .....  \notag \\
& \quad \quad + \gamma_{\eta-1}(w_{i-1}) \bigg  (  \Zb_{\Db_\eta}^{(\rho_\eta, \kappa_\eta)}  (w_i | w_{i-1})  + \gamma_{\eta}(w_{i-1}) \bigg (  \Zb^{(\rho_{\eta+1} = 0, \kappa_{\eta+1} = 1)}  (w_i | w_{i-1})   \bigg  ) \bigg ) ...  \bigg )
\label{eq:overallagain}
\end{flalign}
where we note that $ \Zb^{(\rho_{\eta+1} = 0, \kappa_{\eta+1} = 1)}  (w_i | w_{i-1}) $ is equivalent to $P^\alt(w_i)$. It is assumed that $1 \geq \rho_0 \geq .... \rho_{\eta+1} = 0$.

First unroll the recursion and rewrite  $P_{\plre}(w_i | w_{i-1})$ as:
\begin{flalign}
& P_{\plre}(w_i | w_{i-1}) = \sum_{j=0}^{\eta+1} \gamma_{0:j-1}(w_{i-1}) \Zb_{\Db_j}^{(\rho_j, \kappa_j)}(w_i | w_{i-1}) \notag 
\end{flalign}
where $\gamma_{0:j-1}(w_{i-1}) =  \prod_{h=0}^{j-1} \gamma_h(w_{i-1})$
and $\gamma_{0:-1}(w_{i-1}) = 1$. Note that $P_{\textrm{pwr}}(w_i | w_{i-1})$ can be written in the same way.
\begin{flalign}
& P_{pwr}(w_i | w_{i-1}) =  \sum_{j=0}^{\eta} \gamma_{0:j}(w_{i-1}) \Yb_{\Db_j}^{(\rho_j)}(w_i | w_{i-1}) 
\end{flalign}

Note that $P_{\textrm{pwr}}(w_i | w_{i-1})$ already satisfies the marginal constraint i.e.
\begin{flalign}
& \Phat(w) = \sum_{w_{i-1}} P_{\textrm{pwr}}(w_i | w_{i-1}) \Phat(w_{i-1})
\end{flalign}
because the discounts were chosen such that $P_{\textrm{pwr}}(w_i | w_{i-1}) = \Phat(w_i | w_{i-1})$

Thus it suffices to show that for all $j = 0,...,\eta+1$:
\begin{flalign}
& \sum_{w_{i-1}} \Phat(w_{i-1}) \gamma_{0:j-1} (w_{i-1}) \Yb_{\Db_j}^{(\rho_j)}(w_i | w_{i-1}) =  \sum_{w_{i-1}} \Phat(w_{i-1}) \gamma_{0:j-1} (w_{i-1}) \Zb_{\Db_j}^{(\rho_j, \kappa_j)}(w_i | w_{i-1}) 
\end{flalign}
The statement above is trivially true when $j=0$. For all other cases, note that due to the way we have set the discounts, $\gamma_{0:j-1}$ takes a
special form:

\begin{flalign}
 \prod_{h=0}^{j-1} \gamma_h(w_{i-1}) &= \frac{d_{\ast}
  \sum_{i} c_{i,i-1}^{\rho_{1}} }{\sum_{i} c_{i,i-1}^{\rho_{0}}}  \frac{d_{\ast}
  \sum_{i} c_{i,i-1}^{\rho_{2}} }{\sum_{i} c_{i,i-1}^{\rho_{1}}}...  \frac{d_{\ast}
   \sum_{i} c_{i,i-1}^{\rho_{j}} }{\sum_{i} c_{i,i-1}^{\rho_{j-1}}} \notag \\
    &= \frac{(d_{\ast})^j
   \sum_{i} c_{i,i-1}^{\rho_{j}}}{\sum_{i} c_{i,i-1}}
  \end{flalign}

Using this form in Eq. 5 and simplifying yields:
\begin{eqnarray}
& \sum_{w_{i-1}} \left ( \sum_{i} c_{i,i-1}^{\rho_{j}}  \right )  \Yb_{\Db_j}^{(\rho_j)}(w_i | w_{i-1})  = \sum_{w_{i-1}} \left ( \sum_{i} c_{i,i-1}^{\rho_{j}} \right )    \Zb_{\Db_j}^{(\rho_j, \kappa_j)}(w_i | w_{i-1}) \notag
\end{eqnarray}
which is equivalent to requiring that
\begin{eqnarray}
\sum_{w_{i-1}} \Yb_{\Db_j}^{(\rho_j)}(w_i, w_{i-1})  = \sum_{w_{i-1}} \Zb_{\Db_j}^{(\rho_j, \kappa_j)}(w_i, w_{i-1})
\end{eqnarray}

 which holds because rank minimization under $gKL$ preserves
 row and column sums.  
\end{proof}
\label{lem:marglemma}
\end{lemma}

\section{Generalization to $n > 2$}

\begin{theorem}
Let $P_{\textrm{plre}}(w_i | w^{i-1}_{i-n+1})$ indicate the PLRE smoothed conditional probability and $\Phat(w)$ indicate the MLE probability of $w$. Then,
\begin{flalign}
\Phat(w) = \sum_{w^{i-1}_{i-n+1}} P_{\textrm{plre}}(w_i | w^{i-1}_{i-n+1}) \Phat(w^{i-1}_{i-n+1})
\end{flalign}

\begin{proof}

Recall that,
\begin{flalign}
& P_{\textrm{plre}}(w_i | w^{i-1}_{i-n+1}) = P_{\Db_0}^{\textrm{alt}}(w_i | w^{i-1}_{i-n+1}) \notag \\
&+ \gamma_0(w^{i-1}_{i-n+1}) \bigg ( \Zb_{\Db_1}^{(\rho_1, \kappa_1)} (w_i | w^{i-1}_{i-n+1})  + .....  \notag \\
& + \gamma_{\eta-1}(w^{i-1}_{i-n+1}) \bigg  (  \Zb_{\Db_\eta}^{(\rho_\eta, \kappa_\eta)}  (w_i | w^{i-1}_{i-n+1})  \notag \\
&+ \gamma_{\eta}(w^{i-1}_{i-n+1}) \bigg (  P_{\textrm{plre}}(w_i | w^{i-1}_{i-n+2}) \bigg  ) \bigg ) ...  \bigg )
\label{eq:overallagain}
\end{flalign}

Define,
\begin{flalign}
& P_{\textrm{pwr}}(w_i | w^{i-1}_{i-n+1}) = P_{\Db_0}^{\textrm{alt}}(w_i | w^{i-1}_{i-n+1}) \notag \\
&+ \gamma_0(w^{i-1}_{i-n+1}) \bigg ( \Yb_{\Db_1}^{(\rho_1, \kappa_1)} (w_i | w^{i-1}_{i-n+1})  + .....  \notag \\
& + \gamma_{\eta-1}(w^{i-1}_{i-n+1}) \bigg  (  \Yb_{\Db_\eta}^{(\rho_\eta, \kappa_\eta)}  (w_i | w^{i-1}_{i-n+1})  \notag \\
&+ \gamma_{\eta}(w^{i-1}_{i-n+1}) \bigg (  P_{\textrm{pwr}}(w_i | w^{i-1}_{i-n+2}) \bigg  ) \bigg ) ...  \bigg )
\label{eq:overallpwr}
\end{flalign}
where with a little abuse of notation
\begin{flalign}
 \Yb_{\Db_j}^{\rho_j}(w_i | w^{i-1}_{i-n'+1}) = \frac{\ctil(w_i, w^{i-1}_{i-n'+1})^{\rho_j} - \Db_j(w_i, w^{i-1}_{i-n'+1})}{\sum_{w_i}\ctil(w_i, w^{i-1}_{i-n'+1})^{\rho_j}}
\end{flalign}
and 
\begin{align*}
\hspace{-0.5cm} \ctil(w_i, w^{i-1}_{i-n'+1}) &= \begin{cases} c(w_i, w^{i-1}_{i-n'+1}), &\hspace{-0.3cm} \mbox{if } n' = n \\ \\
 N_{-} (w^{i}_{i-n'+1}) &\hspace{-0.3cm} \mbox{if } n' < n \end{cases}
\end{align*}

Furthermore, define
\begin{flalign}
& P_{\textrm{pwr}}^{\textrm{terms}}(w_i | w^{i-1}_{i-n'+1}) = P_{\Db_0}^{\textrm{alt}}(w_i | w^{i-1}_{i-n'+1}) \notag \\
&\quad + \gamma_0(w^{i-1}_{i-n'+1}) \bigg ( \Yb_{\Db_1}^{(\rho_1, \kappa_1)} (w_i | w^{i-1}_{i-n'+1})  + .....  \notag \\
&\quad  + \gamma_{\eta-1}(w^{i-1}_{i-n'+1}) \bigg  (  \Yb^{(\rho_\eta, \kappa_\eta)}  (w_i | w^{i-1}_{i-n'+1}) \bigg  ) ...  \bigg )
\end{flalign}

Note that because of the way the discounts are computed in Algorithm 1,
\begin{eqnarray}
P_{\textrm{pwr}}^{\textrm{terms}}(w_i | w^{i-1}_{i-n'+1}) = P^{\textrm{alt}}(w_i | w^{i-1}_{i-n'+1})
\end{eqnarray}
for all $n' \leq n$.

As a result, (for some choice of discount)
\begin{eqnarray}
P_{\textrm{pwr}}(w_i | w^{i-1}_{i-n+1}) = P_{\textrm{kn}} (w_i | w^{i-1}_{i-n+1})
\end{eqnarray}

Since, we know that Kneser Ney satisfies the marginal constraint~\cite{Chen1999} this implies that,

\begin{eqnarray}
\Phat(w) = \sum_{w^{i-1}_{i-n+1}} P_{\textrm{pwr}}(w_i | w^{i-1}_{i-n+1}) \Phat(w^{i-1}_{i-n+1})
\end{eqnarray}

Thus, all we have to do is prove that
\begin{eqnarray}
\sum_{w^{i-1}_{i-n+1}} P_{\textrm{pwr}}(w_i | w^{i-1}_{i-n+1}) \Phat(w^{i-1}_{i-n+1}) = \sum_{w^{i-1}_{i-n+1}} P_{\textrm{plre}}(w_i | w^{i-1}_{i-n+1}) \Phat(w^{i-1}_{i-n+1})
\end{eqnarray}

Now, we follow the same argument as with $n=2$ (i.e. unrolling the recursion and applying the fact that $gKL$ preserves row/column sums).

For notational simplicity assume that $n=3$.
Then, we can write $P_{\textrm{pwr}}(w_i | w^{i-1}_{i-n+1})$ as:
\begin{eqnarray}
P_{\textrm{pwr}}(w_i | w^{i-1}_{i-2}) = \sum_{j=0}^{\eta} \gamma_{0:j-1}(w^{i-1}_{i-2}) \Yb_{\Db_j}^{(\rho_j, \kappa_j)} (w_i | w^{i-1}_{i-2}) + \sum_{j=0}^{\eta+1} \gamma_{0:\eta}(w^{i-1}_{i-2}) \gamma_{0:j-1}(w_{i-1}) \Yb_{\Db_j}^{(\rho_j, \kappa_j)} (w_i | w_{i-1})
\end{eqnarray}

where $\gamma_{0:-1}(w^{i-1}_{i-2}) = 1$
and
\begin{flalign}
\gamma_{0:j-1}(w^{i-1}_{i-2}) = \prod_{h=0}^{j-1} \gamma_h(w^{i-1}_{i-2}) &= \frac{d_{\ast}
  \sum_{i} \ctil_{i,i-1,i-2}^{\rho_{1}} }{\sum_{i} \ctil_{i,i-1,i-2}^{\rho_{0}}}  \frac{d_{\ast}
  \sum_{i} \ctil_{i,i-1,i-2}^{\rho_{2}} }{\sum_{i} \ctil_{i,i-1,i-2}^{\rho_{1}}}...  \frac{d_{\ast}
   \sum_{i} \ctil_{i,i-1,i-2}^{\rho_{j}} }{\sum_{i} \ctil_{i,i-1,i-2}^{\rho_{j-1}}} \notag \\
    &= \frac{(d_{\ast})^j
   \sum_{i} \ctil_{i,i-1,i-2}^{\rho_{j}}}{\sum_{i} \ctil_{i,i-1,i-2}}
\end{flalign}
Here $\ctil_{i,i-1,i-2}$ is shorthand for $\ctil(w_i, w^{i-1}_{i-2})$.

Similarly, $\gamma_{0:-1}(w_{i-1}) = 1$ and
\begin{flalign}
\gamma_{0:j-1}(w_{i-1}) = \prod_{h=0}^{j-1} \gamma_h(w_{i-1}) &= \frac{d_{\ast}
  \sum_{i} \ctil_{i,i-1}^{\rho_{1}} }{\sum_{i} \ctil_{i,i-1}^{\rho_{0}}}  \frac{d_{\ast}
  \sum_{i} \ctil_{i,i-1}^{\rho_{2}} }{\sum_{i} \ctil_{i,i-1}^{\rho_{1}}}...  \frac{d_{\ast}
   \sum_{i} \ctil_{i,i-1}^{\rho_{j}} }{\sum_{i} \ctil_{i,i-1}^{\rho_{j-1}}} \notag \\
    &= \frac{(d_{\ast})^j
   \sum_{i} \ctil_{i,i-1}^{\rho_{j}}}{\sum_{i} \ctil_{i,i-1}}
\end{flalign}

(Again, it is assumed that $1 \geq \rho_0 \geq .... \rho_{\eta+1} = 0$.)

Analogously, 

\begin{eqnarray}
P_{\textrm{plre}}(w_i | w^{i-1}_{i-2}) = \sum_{j=0}^{\eta} \gamma_{0:j-1}(w^{i-1}_{i-2}) \Zb_{\Db_j}^{(\rho_j, \kappa_j)} (w_i | w^{i-1}_{i-2}) + \sum_{j=0}^{\eta+1} \gamma_{0:\eta}(w^{i-1}_{i-2}) \gamma_{0:j-1}(w_{i-1}) \Zb_{\Db_j}^{(\rho_h, \kappa_j)} (w_i | w_{i-1})
\end{eqnarray}

Now for any trigram term we prove that
\begin{eqnarray}
\sum_{w^{i-1}_{i-2}}  \gamma_{0:j-1}(w^{i-1}_{i-2}) \Yb_{\Db_j}^{(\rho_j, \kappa_j)} (w_i | w^{i-1}_{i-2}) \Phat(w^{i-1}_{i-2}) = \sum_{w^{i-1}_{i-2}}  \gamma_{0:j-1}(w^{i-1}_{i-2}) \Zb_{\Db_j}^{(\rho_j, \kappa_j)} (w_i | w^{i-1}_{i-2}) \Phat(w^{i-1}_{i-2})
\end{eqnarray}
Plugging in the definition of $\gamma_{0:j-1}$ and simplifying gives

\begin{eqnarray}
\sum_{w^{i-1}_{i-2}}  (    \sum_{i} \ctil_{i,i-1,i-2}^{\rho_{j}} ) \Yb_{\Db_j}^{(\rho_j, \kappa_j)} (w_i | w^{i-1}_{i-2})  = \sum_{w^{i-1}_{i-2}}     ( \sum_{i} \ctil_{i,i-1,i-2}^{\rho_{j}} ) \Zb_{\Db_j}^{(\rho_j, \kappa_j)} (w_i | w^{i-1}_{i-2})
\end{eqnarray}
which is equivalent to 
\begin{eqnarray}
\sum_{w^{i-1}_{i-2}} \Yb_{\Db_j}^{(\rho_j, \kappa_j)} (w_i,  w^{i-1}_{i-2})  = \sum_{w^{i-1}_{i-2}} \Zb_{\Db_j}^{(\rho_j, \kappa_j)} (w_i,  w^{i-1}_{i-2})
\end{eqnarray}
which holds because of the definition of $\Zb$ and the fact that rank minimization under $gKL$ preserves row/column sums.

Now consider any bigram term. We seek to show that:
\begin{flalign}
& \sum_{w^{i-1}_{i-2}}  \gamma_{0:\eta}(w^{i-1}_{i-2}) \gamma_{0:j-1}(w_{i-1}) \Yb_{\Db_j}^{(\rho_j, \kappa_j)} (w_i | w_{i-1}) \Phat(w^{i-1}_{i-2}) \notag \\
& \quad \quad = \sum_{w^{i-1}_{i-2}}  \gamma_{0:\eta}(w^{i-1}_{i-2}) \gamma_{0:j-1}(w_{i-1}) \Zb_{\Db_j}^{(\rho_j, \kappa_j)} (w_i | w_{i-1}) \Phat(w^{i-1}_{i-2})
\end{flalign}
Substituting definition of  $\gamma_{0:\eta}(w^{i-1}_{i-2})$ gives
\begin{flalign}
& \sum_{w^{i-1}_{i-2}}\frac{(d_{\ast})^{\eta+1}
   \sum_{i} \ctil_{i,i-1,i-2}^{\rho_{\eta+1}}}{\sum_{i} \ctil_{i,i-1,i-2}} \gamma_{0:j-1}(w_{i-1}) \Yb_{\Db_j}^{(\rho_j, \kappa_j)} (w_i | w_{i-1}) \Phat(w^{i-1}_{i-2}) \notag \\
& \quad \quad   = \sum_{w^{i-1}_{i-2}}  \frac{(d_{\ast})^{\eta+1}
      \sum_{i} \ctil_{i,i-1,i-2}^{\rho_{\eta+1}}}{\sum_{i} \ctil_{i,i-1,i-2}}  \gamma_{0:j-1}(w_{i-1}) \Zb_{\Db_j}^{(\rho_j, \kappa_j)} (w_i | w_{i-1}) \Phat(w^{i-1}_{i-2})
\end{flalign}

Simplifying and pushing in the sum over $w_{i-2}$ gives,
\begin{eqnarray}
\sum_{w_{i-1}}
   (\sum_{i,i-2} \ctil_{i,i-1,i-2}^{\rho_{\eta+1}}) \gamma_{0:j-1}(w_{i-1}) \Yb_{\Db_j}^{(\rho_j, \kappa_j)} (w_i | w_{i-1}) = \sum_{w_{i-1}}  
      (\sum_{i,i-2} \ctil_{i,i-1,i-2}^{\rho_{\eta+1}}) \gamma_{0:j-1}(w_{i-1}) \Zb_{\Db_j}^{(\rho_j, \kappa_j)} (w_i | w_{i-1})
\end{eqnarray}
Note that since $\rho_{\eta+1} = 0$,
$\sum_{i,i-2} \ctil_{i,i-1,i-2}^{\rho_{\eta+1} = 0} = \sum_{i} \ctil_{i,i-1}$ (by definition of $\ctil$).

Using this fact and substituting definition of  $\gamma_{0:j-1}(w_{i-1})$ gives
\begin{eqnarray}
\sum_{w_{i-1}}
   (\sum_{i} \ctil_{i,i-1}) \frac{(d_{\ast})^j
      \sum_{i} \ctil_{i,i-1}^{\rho_{j}}}{\sum_{i} \ctil_{i,i-1}} \Yb_{\Db_j}^{(\rho_j, \kappa_j)} (w_i | w_{i-1}) = \sum_{w_{i-1}}  
      (\sum_{i} \ctil_{i,i-1}) \frac{(d_{\ast})^j
         \sum_{i} \ctil_{i,i-1}^{\rho_{j}}}{\sum_{i} \ctil_{i,i-1}} \Zb_{\Db_j}^{(\rho_j, \kappa_j)} (w_i | w_{i-1})
\end{eqnarray}

Simplifying gives,
\begin{eqnarray}
\sum_{w_{i-1}} \Yb_{\Db_j}^{(\rho_j)}(w_i, w_{i-1})  = \sum_{w_{i-1}} \Zb_{\Db_j}^{(\rho_j, \kappa_j)}(w_i, w_{i-1})
\end{eqnarray}

 which holds because rank minimization under KL divergence preserves
 row and column sums.


\end{proof}
\end{theorem}

\bibliographystyle{acl}
\bibliography{thesis-bib,tacl2013}
